\documentclass{article}
\newcommand{\hdrTitle}[1]{\textbf{#1}}

\usepackage[preprint]{neurips_2026} 
\usepackage{amsmath}
\newcommand{\hdrSub}[1]{\textit{#1}}

\usepackage[utf8]{inputenc} % allow utf-8 input
\usepackage[T1]{fontenc}    % use 8-bit T1 fonts
\usepackage{hyperref}       % hyperlinks
\usepackage{url}            % simple URL typesetting
\usepackage{booktabs}       % professional-quality tables
\usepackage{amsfonts}       % blackboard math symbols
\usepackage{nicefrac}       % compact symbols for 1/2, etc.
\usepackage{microtype}      % microtypography
\usepackage{float}
\usepackage{graphicx}
\usepackage{todonotes}
\usepackage{tcolorbox}
\tcbuselibrary{raster}
\usepackage{tikz}
\usetikzlibrary{positioning, arrows.meta, calc}
\usetikzlibrary{shadows, shadings}
\usetikzlibrary{shapes.geometric}

\definecolor{peAccent}{HTML}{565660}
\definecolor{dpoAccent}{HTML}{3D5A7B}
\definecolor{specAccent}{HTML}{2A7268}
\definecolor{peFill}{HTML}{EEEEF1}
\definecolor{dpoFill}{HTML}{E3EBF3}
\definecolor{specFill}{HTML}{DDEEEB}
\definecolor{specBg}{HTML}{F7FBFA}

\definecolor{peHdr}{HTML}{2A4A7A}
\definecolor{peBoxT}{HTML}{6A85B6}
\definecolor{peBoxB}{HTML}{4A6A9A}
\definecolor{dpoHdr}{HTML}{2C7DB8}
\definecolor{dpoBoxT}{HTML}{72B5E6}
\definecolor{dpoBoxB}{HTML}{4F9CCF}
\definecolor{specHdr}{HTML}{2E8B57}
\definecolor{specBoxT}{HTML}{6FBD92}
\definecolor{specBoxB}{HTML}{4FA678}
\definecolor{specBgPanel}{HTML}{EAF7EE}
\definecolor{specPhaseTint}{HTML}{F5FBF7}
\definecolor{titleBar}{HTML}{1F1F2E}
\definecolor{annotGray}{HTML}{687283}

\definecolor{procTop}{HTML}{6BA092}
\definecolor{procBot}{HTML}{4A7C6D}
\definecolor{ioTop}{HTML}{B8A99A}
\definecolor{ioBot}{HTML}{94867A}
\definecolor{decTop}{HTML}{D4B58A}
\definecolor{decBot}{HTML}{A88860}

\title{Towards Spec Learning: Inference-Time Alignment from Preference Pairs}

\author{
  Dhriti Krishnan\thanks{Equal contribution.}\quad
  Tejas Goyal\thanks{Equal contribution.}\quad
  Jaromir Savelka \\
  Department of Computer Science \\
  Carnegie Mellon University \\
  \texttt{\{dhritik, tejasgoy, jsavelka\}@andrew.cmu.edu}
}

\begin{document}

\maketitle

\begin{abstract}
 Steering a large language model (LLM) toward a desired behavior typically relies on an iterative process of hand-crafting a prompt based on a careful inspection of the model's responses. This is an involved, brittle, and error-prone process. Preference-based fine-tuning is a more rigorous but often prohibitively expensive solution. We propose \emph{spec learning}, a framework that relies on a brief user instruction and a small set of preference judgments. These are compiled into specifications in the form of natural-language prompts for an LLM. Specifications condition LLMs at inference time, and no parameter updates to the underlying models are required. We show that the responses generated based on the compiled specifications often outperform direct preference optimization (DPO) on datasets from specialized domains whose preference signal is dense. Unlike opaque weight updates, the resulting specifications are human-readable and double as interpretable and transparent written embodiments of the preference signal that produced them.
  % \todo[inline]{We need to ensure the abstract and/or introduction clearly meets the following criteria: The abstract and/or introduction should clearly state the claims made, including the contributions made in the paper and important assumptions and limitations. The claims made should match theoretical and experimental results, and reflect how much the results can be expected to generalize to other settings. It is fine to include aspirational goals as motivation as long as it is clear that these goals are not attained by the paper. }
\end{abstract}

\section{Introduction}

% \begin{figure}[!b]
%     \centering
%     \includegraphics[width=1\linewidth]{images/main_diagram.png}
%     \caption{Caption}
%     \label{fig:concept}
% \end{figure}

\definecolor{peAccent}{HTML}{565660}
\definecolor{dpoAccent}{HTML}{3D5A7B}
\definecolor{specAccent}{HTML}{2A7268}
\definecolor{peFill}{HTML}{EEEEF1}
\definecolor{dpoFill}{HTML}{E3EBF3}
\definecolor{specFill}{HTML}{DDEEEB}
\definecolor{specBg}{HTML}{F7FBFA}

\begin{figure}[!t]
\centering
\begin{tikzpicture}[
  font=\sffamily\footnotesize,
  >={Latex[length=1.7mm, width=1.4mm]},
  hdrStyle/.style={
    text=white,
    minimum width=4.3cm, text width=3.95cm,
    minimum height=11mm, rounded corners=2pt,
    align=center, inner sep=2.5pt,
    drop shadow={shadow xshift=0.5mm, shadow yshift=-0.5mm,
                 opacity=0.30, fill=black!70},
  },
  hdrTitle/.style={font=\sffamily\scriptsize\bfseries},
  hdrSub/.style={font=\sffamily\tiny\itshape, text=white!92},
  peBox/.style={
    minimum width=2.4cm, minimum height=7.5mm,
    top color=peBoxT, bottom color=peBoxB,
    text=white, draw=peBoxB, line width=0.5pt,
    rounded corners=2pt, align=center, inner sep=1.5pt,
    font=\sffamily\scriptsize\bfseries,
    drop shadow={shadow xshift=0.4mm, shadow yshift=-0.4mm,
                 opacity=0.25, fill=black!60},
  },
  dpoBox/.style={
    minimum width=2.0cm, minimum height=7.5mm,
    top color=dpoBoxT, bottom color=dpoBoxB,
    text=white, draw=dpoBoxB, line width=0.5pt,
    rounded corners=2pt, align=center, inner sep=1.5pt,
    font=\sffamily\scriptsize\bfseries,
    drop shadow={shadow xshift=0.4mm, shadow yshift=-0.4mm,
                 opacity=0.25, fill=black!60},
  },
  specBox/.style={
    minimum width=1.25cm, minimum height=6.5mm,
    top color=specBoxT, bottom color=specBoxB,
    text=white, draw=specBoxB, line width=0.5pt,
    rounded corners=2pt, align=center, inner sep=1pt,
    font=\sffamily\tiny\bfseries,
    drop shadow={shadow xshift=0.3mm, shadow yshift=-0.3mm,
                 opacity=0.22, fill=black!60},
  },
  phaseBox/.style={
    rounded corners=3pt, fill=specPhaseTint,
    draw=specHdr!50, line width=0.4pt, dashed,
  },
  phaseLbl/.style={font=\sffamily\tiny\bfseries, text=specHdr},
  annot/.style={
    font=\sffamily\tiny\itshape, text=annotGray,
    align=center,
  },
  arr/.style={->, line width=0.55pt, draw=black!60},
  loopArr/.style={->, line width=0.55pt, draw=black!60, rounded corners=4pt},
]

\def\colA{0}
\def\colB{4.7}
\def\colC{9.4}

% ============================================================
% PANEL 1 — Manual Prompt Engineering
% ============================================================
\node[hdrStyle, fill=peHdr] at (\colA, 0)
  {{\hdrTitle MANUAL PROMPT\\ENGINEERING}\\[0.3ex]
   {\hdrSub Signal in User}};

\node[peBox] (p1user) at (\colA, -1.5) {User writes\\prompt};
\node[peBox] (p1llm)  at (\colA, -3.0) {LLM};
\node[peBox] (p1resp) at (\colA, -4.5) {Response};

\draw[arr] (p1user) -- (p1llm);
\draw[arr] (p1llm)  -- (p1resp);

\draw[loopArr]
  (p1resp.west) -- ++(-0.5, 0)
                |- (p1user.west);

\node[annot, anchor=north] at (\colA, -5.05)
  {no preferences\\codified};

% ============================================================
% PANEL 2 — Preference Fine-Tuning (DPO)
% ============================================================
\node[hdrStyle, fill=dpoHdr] at (\colB, 0)
  {{\hdrTitle PREFERENCE FINE-TUNING\\(DPO)}\\[0.3ex]
   {\hdrSub Signal in Weights}};

\foreach \i in {0, 1, 2, 3} {
  \fill[dpoBoxT!55, draw=dpoBoxB, line width=0.3pt, rounded corners=1pt]
    ($(\colB - 1.5, -1.5) + (\i*0.07, -\i*0.07) - (0.4, 0.32)$) rectangle
    ($(\colB - 1.5, -1.5) + (\i*0.07, -\i*0.07) + (0.4, 0.32)$);
}
\node[annot, anchor=north] at (\colB - 1.5, -2.05)
  {large pref dataset\\(N\,=\,1{,}000)};

\node[dpoBox] (p2ft)   at (\colB + 0.7, -1.5) {Fine-Tuning\\\scriptsize (LoRA)};
\node[dpoBox] (p2llm)  at (\colB + 0.7, -3.0) {Fine-Tuned LLM\\\scriptsize $\mathrm{LLM}_{\theta'}$};
\node[dpoBox] (p2resp) at (\colB + 0.7, -4.5) {Response};

\draw[arr] (\colB - 0.85, -1.5) -- (p2ft.west);
\draw[arr] (p2ft)  -- (p2llm);
\draw[arr] (p2llm) -- (p2resp);

\node[annot, anchor=north] at (\colB + 0.7, -5.05)
  {whole model\\specialized};

% ============================================================
% PANEL 3 — Spec Learning
% ============================================================
\node[hdrStyle, fill=specHdr] at (\colC, 0)
  {{\hdrTitle SPEC LEARNING}\\[0.3ex]
   {\hdrSub Signal in Input}};

\node[phaseBox, minimum width=4.4cm, minimum height=2.5cm,
      anchor=north] (phase1) at (\colC, -0.85) {};
\node[phaseLbl, anchor=north west, inner sep=2.5pt]
  at (phase1.north west)
  {Phase 1: Compile};

\node[specBox] (brief)    at (\colC - 1.4, -1.75) {Brief\\Instruction};
\node[specBox] (pref)     at (\colC - 1.4, -2.75) {Pref Pairs\\(N\,=\,20)};
\node[specBox] (compiler) at (\colC,        -2.25) {LLM\\Compiler};
\node[specBox] (specs)    at (\colC + 1.4, -2.25) {Specs};

\draw[arr] (brief.east)    -- (compiler.west);
\draw[arr] (pref.east)     -- (compiler.west);
\draw[arr] (compiler.east) -- (specs.west);

\node[phaseBox, minimum width=4.4cm, minimum height=1.4cm,
      anchor=north] (phase2) at (\colC, -3.85) {};
\node[phaseLbl, anchor=north west, inner sep=2.5pt]
  at (phase2.north west)
  {Phase 2: Apply};

\node[specBox] (uquery) at (\colC - 1.4, -4.85) {User Query};
\node[specBox] (sllm)   at (\colC,        -4.85) {LLM\\(unchanged)};
\node[specBox] (fresp)  at (\colC + 1.4, -4.85) {Final\\Response};

\draw[arr] (uquery) -- (sllm);
\draw[arr] (sllm)   -- (fresp);

\draw[arr, rounded corners=3pt]
  (specs.south) -- ++(0, -0.925)
                -| (sllm.north);

\node[annot, anchor=north] at (\colC, -5.60)
  {transparent embodiment\\of preference signal};

% Vertical dividers between panels
\draw[draw=black!22, line width=0.4pt, dashed] (2.35, 0.55) -- (2.35, -6.30);
\draw[draw=black!22, line width=0.4pt, dashed] (7.05, 0.55) -- (7.05, -6.30);

\end{tikzpicture}
\caption{Three regimes for shaping LLM behavior, distinguished by where the alignment signal lands. \emph{Manual prompt engineering} relies on the user to iterate prompts and inspect responses; no preference data is captured. \emph{Preference fine-tuning} (DPO) consumes $\sim$1{,}000 preference pairs to specialize the model weights. \emph{Spec learning} (ours) compiles 20 preference judgments into a natural-language system prompt offline; at inference the prompt is composed with the user query and consumed by the same unchanged base model.}
\label{fig:concept}
\end{figure}
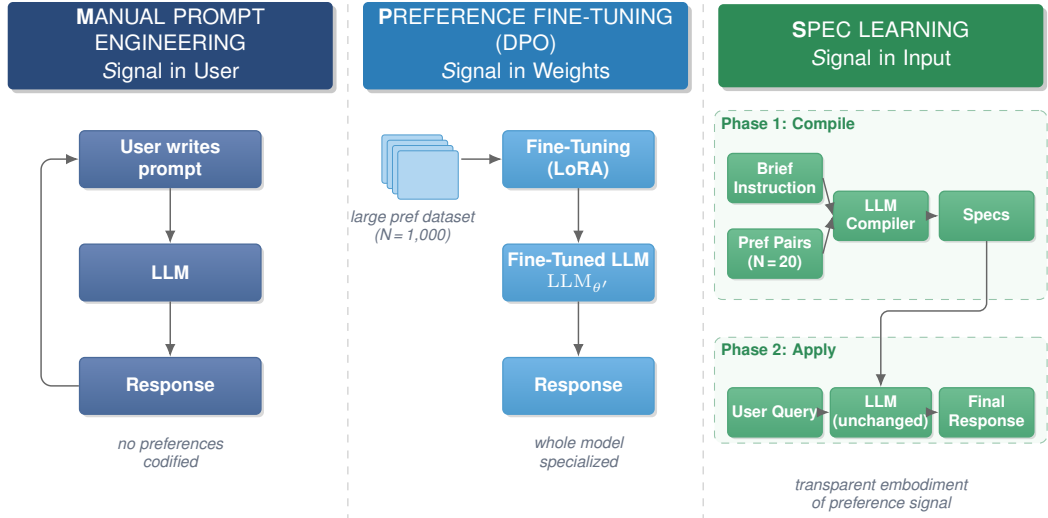

We explore the emerging trend of streamlining users' interactions with large language models~(LLMs) by automatically generating well-performing prompts from low-effort user input. %Historically, there have been several dimensions which have been of central importance to AI research such as, e.g., model performance, efficiency, or explainability. One such dimension is the \emph{mode in which AI can be leveraged} and integrated into useful systems. While earlier ML approaches have been dependent on domain experts for meticulous feature engineering, deep learning provided a viable alternative by automating the tedious process. The approach of first pre-training a model on a general weakly supervised objective and subsequently fine-tuning it on a downstream task enabled the application of powerful data hungry methods in domains and tasks where data have been sparse. Despite all these advances the integration of AI into real-world systems required highly specialized ML engineering expertise. 
The recent proliferation of zero-shot prompting~\cite{liu2023pre} and in-context learning \cite{brown2020language,dong2024survey,mueller2024context} enabled much wider range of users and builders to leverage AI in their systems and workflows. At the same time, \emph{prompt engineering} has been coined as the new term since constructing a well-performing prompt and/or selecting the right examples to be included in the prompt turned out to be involved and brittle process~\cite{gonen2023demystifying,zhao2021calibrate,min2022rethinking}. Many studies have shown that to prompt LLMs effectively may require specialized expertise and in many cases cannot be performed reliably by non-professionals~\cite{zamfirescu2023johnny}. Hence, relieving the user from much of the onus of prompt engineering has been gaining in prominence and has established itself as one of the most important frontiers in present day AI research~\cite{ramnath2025systematic,chiang2023can}.

We propose \emph{spec learning}, a framework in which a brief user instruction and a small set of preference judgments are compiled into a system prompt that conditions an LLM at inference. Conventional prompt engineering leaves three steps to the user: writing the initial prompt, iterating on it, and judging whether the output is good. Most users execute this loop poorly~\cite{zamfirescu2023johnny}. They settle for weak prompts, skip evaluation, and in higher-stakes settings risk being misled by the model. Spec learning splits the labor differently. The user supplies only the inputs needed to express intent, a brief instruction and a small set of preference judgments, and the proposer, judge, and synthesizer LLMs produce the prompt (Figure~\ref{fig:concept}).

Our central claim is that around 20 preference pairs already contain enough signal to compile a natural language prompt that matches gradient-based preference tuning trained on a corpus fifty times larger, while leaving the inference model's weights unchanged. The claim depends on three fundamental assumptions: (i) the preference signal in a target domain admits a compact set of explicit principles, (ii) that the proposer and judge LLMs are strong enough to find and verify those principles from limited data, and (iii) that the inference model can act on detailed natural language instructions. The assumptions explain where our method breaks. Preferences that span heterogeneous tasks under a single label resist compression into a unified rule set, and we observe this on Anthropic's HH-RLHF helpful subset \cite{bai2022training}, where the compiled spec only marginally improves over random chance ($0.55$) while DPO captures a real preference signal ($0.70$)~(Section~\ref{sec:results-headline}). When the underlying signal is rich and does not lend itself to be expressed in a small set of rules, gradient-based methods recover what spec learning cannot. A practical caveat is that the LLM judge used to validate the proposed framework may itself be biased along length, position, and model family lines~\cite{wang_large_2024,chen_humans_2024,haldar_rating_2025}. We calibrate against gold preferences and swap response positions. Spec learning is not meant to replace preference tuning when thousands of labels are available. It is rather suitable for situations where the user would otherwise have to rely on unprincipled prompt engineering.

\section{Background and Related Work}

% \subsection{Prompt Induction and Optimization}
There has been a considerable interest in automatic prompt optimization (APO) over the recent years taking several distinct approaches. Evolutionary-based algorithms maintain a population of prompt candidates, apply mutation/crossover via an LLM or hand-coded operators, and select survivors based on validation-set fitness \cite{xu_gps_2022,fernando_promptbreeder_2023,guo_evoprompt_2025,agrawal_gepa_2026,zhang_sprig_2026,sinha_survival_2024,sun_fedbpt_2023,pan_plum_2024}. Reinforcement learning-based approaches cast prompt construction or editing as a sequential decision problem and train~(or query) a policy with task-derived reward \cite{diao_black-box_2023,dong_pace_2024,kong_prewrite_2024,jafari_morl-prompt_2024,sun_query-dependent_2024,zhang_tempera_2023}. Gradient-based methods update prompts using the actual or natural-language gradients \cite{shin_autoprompt_2020,shi_auto-prompt_2025,yuksekgonul_textgrad_2024}. An interesting group of methods inspect failure and success cases, generate natural language critiques, and rewrite the prompt~\cite{pryzant_automatic_2023,he_crispo_2025,juneja_task_2025,sun_autohint_2023,wu_strago_2024,yang_ampo_2024,zhang_prefer_2024}. A similar group of methods is based on LLM being conditioned on prior prompts and their scores, and asked to propose better prompts without an explicit critique~\cite{yang_large_2024,zhou_large_2023,ye_prompt_2024,yang_dual-phase_2024}. Another popular approach is to search the discrete prompt space via local edits, or planning algorithms~\cite{prasad_grips_2023,hsieh_automatic_2024,schnabel_symbolic_2024,wang_promptagent_2023,zhan_unveiling_2024}. Preference-based optimization learns from preference comparisons to steer prompts toward aligned outputs \cite{cheng_black-box_2024,jin_apeer_2025,trivedi_align-pro_2025}. As agentic systems gain prominence approaches that optimize prompts jointly across multiple LLM calls or stages of an LLM agent have gained increased attention~\cite{opsahl-ong_optimizing_2024,sordoni_joint_2023,chen_prompt_2024}. Further, there have been many more attempts on APO such as, e.g., InstructZero \cite{chen_instructzero_2023}, INSTINCT \cite{lin_use_2024}, FIPO \cite{lu_fipo_2025}, MoP \cite{wang_one_2024}, Self-Instruct \cite{wang_self-instruct_2023}, GPO \cite{li_robust_2023}, DialPrompt \cite{liu_taming_2025}, or IntentGPT \cite{rodriguez_intentgpt_2024}. 

% All the above mentioned methods require non-trivial resources to be created upfront, most often in the form of labeled data points (between higher tens and many thousands). Hence, their employment in spec learning is problematic as the core goal is to offload as much effort from the user as possible~(i.e., a brief prompt and limited amount of feedback). Prior work has demonstrated that LLMs can explicitly infer an underlying task from a few demonstrations \cite{honovich_instruction_2023,choi_hard_2024,xian_prompts_2024,deng_rlprompt_2024,zhou_survival_2023,xu_reprompting_2024,wan_teach_2024} or preference data \cite{lin_prompt_2024}. Similarly, it has been shown that prompts can be successfully optimized through limited amount of linguistic feedback (critique) \cite{shinn_reflexion_2023,agarwal_promptwizard_2024,long_prompt_2024}. 
% Interestingly, APO with a limited amount of user input attracted a lot of attention in computer vision research, specifically for text-to-image methods \cite{du_ipo_2024,liu_language_2024,manas_improving_2024,mirza_glov_2025}. Inverse Constitutional AI is a sample-efficient approach to extract interpretable principles from preference feedback datasets via compression and reconstruction \cite{findeis_inverse_2025}. ConstitutionMaker is an interactive tool that converts user feedback into natural language principles for steering LLM outputs via prompt updates that requires no training data, only user feedback in interactive sessions \cite{petridis_constitutionmaker_2024}. 

All the above mentioned methods require non-trivial resources to be created upfront, most often in the form of labeled data points (between higher tens and many thousands). Hence, their employment in spec learning is problematic as the core goal is to offload as much effort from the user as possible~(i.e., at most a brief instruction and a small set of preferences). Prior work has demonstrated that LLMs can explicitly infer an underlying task from a few demonstrations \cite{honovich_instruction_2023,choi_hard_2024,xian_prompts_2024,deng_rlprompt_2024,zhou_survival_2023,xu_reprompting_2024,wan_teach_2024} or preference data \cite{lin_prompt_2024}. Similarly, it has been shown that prompts can be successfully optimized through limited amount of linguistic feedback (critique) \cite{shinn_reflexion_2023,agarwal_promptwizard_2024,long_prompt_2024}. A body of work on text-to-image prompt induction with limited user input applies analogous techniques  \cite{du_ipo_2024,liu_language_2024,manas_improving_2024,mirza_glov_2025}.

The most closely related prior work is Inverse Constitutional AI \cite{findeis_inverse_2025}. ICAI extracts interpretable principles from preference data through a compression and reconstruction objective, producing a constitution that summarizes the dataset. Our compiler adapts the ICAI principle-induction pipeline with several modifications described in Section~\ref{sec:preliminary} (swap-and-average validation step, composite ranking that combines prevalence and accuracy, and the Janus synthesizer~\cite{lee_aligning_2024} for register-controlled prompt assembly). %We do not invoke the released ICAI codebase directly. 
Further, ConstitutionMaker \cite{petridis_constitutionmaker_2024} converts interactive user feedback into natural-language principles for prompt steering. This is relevant for spec learning as we primarily envision the framework to be used interactively.

There is a consistent line of work focused on modifying model weights directly using preference data. Reinforcement learning from human feedback trains a reward model from pairwise judgments and uses policy gradients to align an LLM with the resulting reward signal \cite{christiano_deep_2017,stiennon_learning_2020,ouyang_training_2022,bai_training_2022}. Direct Preference Optimization bypasses the reward model by reparameterizing the alignment objective into a closed-form contrastive loss \cite{rafailov_direct_2023}, and a family of related contrastive objectives has since extended this approach \cite{azar_general_2023,ethayarajh_kto_2024,hong_orpo_2024,meng_simpo_2024}. Parameter-efficient adapters such as LoRA \cite{hu_lora_2022} make these methods practical at scale, but they still require thousands of labeled pairs and full backward passes through the policy. Spec learning relies on similar supervision signal but applies it at inference-time in the form of a prompt rather than as a gradient update.

% \subsection{Automated Evaluation}
LLM-as-a-Judge has recently emerged as a scalable evaluation paradigm \cite{gu_survey_2026,zheng_judging_2023}. Multiple methods have been proposed to produce a dedicated LLM as a judge such as, e.g., CritiqueLM~\cite{ke_critiquellm_2024}, self-taught evaluators \cite{wang_self-taught_2024}, JudgeLM \cite{zhu_judgelm_2025}, RL-based J1 \cite{whitehouse_j1_2025}, Think-J \cite{huang_think-j_nodate}, or SFT warm-up with DPO enhancement \cite{yu_improve_nodate}. A number of inference-time techniques to improve judge outputs have been described as well: calibration with provable guarantees \cite{jung_trust_2024}, verifiable global explanations~\cite{gajcin_interpreting_2025}, multi-judge aggregation \cite{badshah_reference-guided_nodate}, or pairwise framing \cite{liu_aligning_2025}. LLM-as-a-judge approach has been successfully utilized in multilingual \cite{son_llm-as--judge_2024} and multimodal \cite{chen_mllm-as--judge_nodate} settings, in multi-step agentic workflows~\cite{zhuge_agent-as--judge_2024}, as well as for code understanding \cite{zhao_codejudge-eval_nodate}. At the same time, it has been shown that LLM judges are vulnerable to prompt-based attacks exploiting various biases (e.g., gender, authority, or confidence)~\cite{chen_humans_2024,ye_justice_2024,wang_large_2024,lee_are_nodate}. Additionally, LLM judges show high variance in ratings across multiple runs on the same prompt \cite{haldar_rating_2025}. It also appears that only the largest models show reasonable human alignment \cite{thakur_judging_2025}. It has been also demonstrated that universal adversarial phrases appended to responses can deceive judges into assigning false positive or inflated scores \cite{raina_is_2024,zhao_one_2025}. Nevertheless, LLM-as-a-judge has been established as an important and promising paradigm and there are now numerous benchmarks focused on measuring judge quality \cite{tan_judgebench_2025,bavaresco_llms_2025,wang_dhp_nodate,raju_constructing_2024,lan_criticeval_nodate,bai_benchmarking_nodate}. We envision and evaluate using LLM-as-a-judge in spec learning to validate the quality of the output.

\section{Data}
\label{sec:setup}

\label{sec:data}
\begin{table}
  \caption{Preference datasets used in our experiments. Pair counts reflect the corpus prior to our content-hashed train/test split. Stack Exchange is streamed and capped during compile rather than fully materialized.}
  \label{tab:datasets}
  \centering
  \small
  \begin{tabular}{lll}
    \toprule
    Dataset           & Domain                                    & Pairs available \\
    \midrule
    Math-DPO          & Multi-step math reasoning                 & 2{,}418         \\
    Code-Pref         & Code generation with transplanted bugs    & 54{,}024        \\
    Code-Security     & Secure vs.\ vulnerable code               & 4{,}656         \\
    Stack-Exchange    & Open-domain technical Q\&A                & 26.8M           \\
    PsyCoPref         & Therapy responses, clinical ratings       & 36{,}653        \\
    Truthy-DPO        & Factual self-acknowledgment  & 1{,}016         \\
    HH-Helpful        & General-purpose helpfulness               & {$\sim$}43{,}000 \\
    \bottomrule
  \end{tabular}
\end{table}

We evaluate spec learning on seven preference datasets (Table~\ref{tab:datasets}). They were chosen to cover a range from technical tasks, where the better response is largely a matter of correctness, to open-ended tasks, where preference depends on tone, structure, or judgment. The technical end includes math reasoning \cite{argilla_distilabel_math_2024}, code correctness \cite{vezora_code_pairs_2024}, code security \cite{cybernative_code_security_2024}, and Stack Exchange Q\&A \cite{lvwerra_stackexchange_2023}. Two judgment-heavy datasets sit in the middle: psychotherapy responses rated on clinical features \cite{psychotherapy_llm_psycopref_2024} and factual humility against confident misconceptions \cite{durbin_truthy_2023}. The open end is Anthropic's HH-RLHF helpful subset \cite{bai_training_2022}, whose preferences span arbitrary user requests. This spread lets us test whether the method depends on the preference being expressible as a small set of rules.

Compile, DPO training, and evaluation each draw from the dataset independently, so a non-deterministic train/test split would risk an evaluation pair leaking into the compile or training arm. To rule this out, every pair is assigned to a partition by a hash of its contents (instruction, chosen, rejected). The same pair therefore always lands in the same split regardless of row order, sampling seed, or downstream filter, and the arms do not need to coordinate. One dataset cannot supply the standard budget. Truthy-DPO contains $1{,}016$ pairs in total, so we cap DPO training at $N=900$ on that dataset and shrink the test partition accordingly.

% We score every comparison with two judges drawn from disjoint model families to mitigate family-aligned reward bias \cite{panickssery_2024_llm}. The primary judge is Gemma 4 31B, and the secondary judge is Kimi K2.6 \cite{moonshotai_kimi_k26_2026}. We report each judge's verdicts separately and flag datasets on which the two diverge.

% We use a single primary judge across all evaluations: GLM-5.1, a frontier mixture-of-experts model whose family is disjoint from both the proposer and the policy under test, mitigating family-aligned reward bias \cite{panickssery_2024_llm}. We selected GLM-5.1 after calibrating candidate judges against gold preference labels on held-out pairs from Math-DPO, Truthy-DPO, and Stack-Exchange: GLM-5.1 recovered the gold label on 60--70\% of pairs across datasets, while a closed-source frontier alternative collapsed to chance on Math-DPO and was disqualified.

\section{Spec Learning Framework}
\label{sec:preliminary}

We summarize the pipeline configuration as a tuple $(\mathcal{D}_N, \sigma, \mathcal{S}, \mathcal{P})$ where $\mathcal{D}_N$ is the set of $N$ preference pairs, $\sigma$ is the selection strategy, $\mathcal{S}$ is the synthesizer, and $\mathcal{P}$ is the proposer. In order to support several important design decisions we performed a number of preliminary experiments to identify the best configurations at each stage.  Each ablation varies one element of this tuple while holding the others at defaults ($N{=}20$, $\sigma{=}$\textsc{random}, $\mathcal{S}{=}$\textsc{janus}, $\mathcal{P}{=}$Gemma 4 31B).

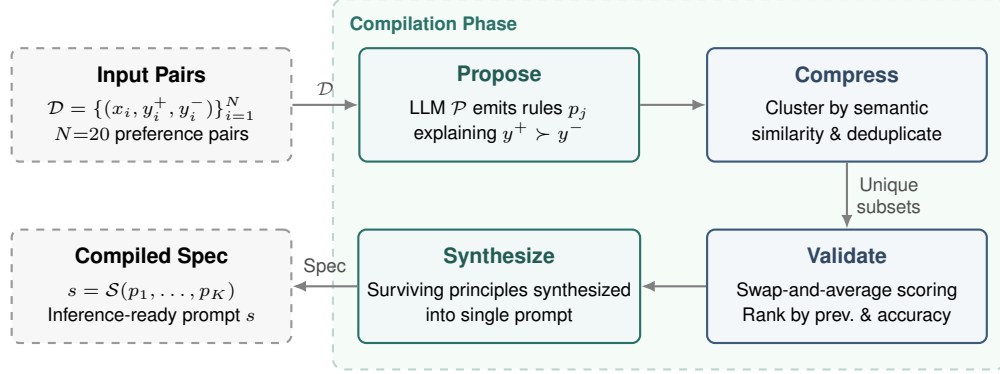
\begin{figure}[!t]
\centering
\begin{tikzpicture}[
  font=\sffamily\footnotesize,
  >={Latex[length=1.8mm, width=1.5mm]},
  % Node styles
  node box/.style={
    rectangle, rounded corners=3pt, thick, align=center,
    minimum width=3.7cm, minimum height=1.5cm
  },
  data/.style={node box, draw=black!40, fill=black!03, dashed},
  llm/.style={node box, draw=specAccent, fill=specFill!40},
  alg/.style={node box, draw=dpoAccent, fill=dpoFill!40},
  % Arrow style
  arr/.style={->, thick, draw=black!50, rounded corners=4pt},
  lbl/.style={font=\sffamily\scriptsize, text=black!70, align=center}
]

% --- Background Phase Box ---
\filldraw[fill=specPhaseTint, draw=specAccent!30, dashed, thick, rounded corners=4pt]
    (2.4, 1.4) rectangle (11.3, -3.5);
\node[anchor=north west, font=\sffamily\scriptsize\bfseries, text=specAccent] 
    at (2.5, 1.3) {Compilation Phase};

% --- Nodes ---
% Row 1
\node[data] (in)   at (0, 0)     {\textbf{Input Pairs}\\[0.1cm] \scriptsize $\mathcal{D} = \{(x_i, y^+_i, y^-_i)\}_{i=1}^{N}$\\ \scriptsize $N{=}20$ preference pairs};

\node[llm]  (prop) at (4.6, 0)   {\textcolor{specAccent!80!black}{\textbf{Propose}}\\[0.1cm] \scriptsize LLM $\mathcal{P}$ emits rules $p_j$\\ \scriptsize explaining $y^+ \succ y^-$};

\node[alg]  (comp) at (9.2, 0)   {\textcolor{dpoAccent!80!black}{\textbf{Compress}}\\[0.1cm] \scriptsize Cluster by semantic\\ \scriptsize similarity \& deduplicate};

% Row 2
\node[alg]  (val)  at (9.2, -2.4) {\textcolor{dpoAccent!80!black}{\textbf{Validate}}\\[0.1cm] \scriptsize Swap-and-average scoring\\ \scriptsize Rank by prev.\ \& accuracy};

\node[llm]  (syn)  at (4.6, -2.4) {\textcolor{specAccent!80!black}{\textbf{Synthesize}}\\[0.1cm] \scriptsize Surviving principles synthesized\\ \scriptsize into single prompt};

\node[data] (out)  at (0, -2.4)   {\textbf{Compiled Spec}\\[0.1cm] \scriptsize $s = \mathcal{S}(p_1, \ldots, p_K)$\\ \scriptsize Inference-ready prompt $s$};

% --- Arrows & Labels ---
\draw[arr] (in.east) -- node[lbl, above] {$\mathcal{D}$} (prop.west);
\draw[arr] (prop.east) -- node[lbl, above] {} (comp.west);
\draw[arr] (comp.south) -- node[lbl, right] {Unique\\subsets} (val.north);
\draw[arr] (val.west) -- node[lbl, above] {} (syn.east);
\draw[arr] (syn.west) -- node[lbl, above] {Spec} (out.east);

\end{tikzpicture}
\caption{Spec compilation pipeline. The framework brackets heavy generative stages (\textsc{propose}, \textsc{synthesize}) with fast, deterministic algorithmic stages (\textsc{compress}, \textsc{validate}) to extract robust principles. The entire compilation operates offline, and the resulting artifact $s$ steers the model without updating its weights.}
\label{fig:pipeline}
\end{figure}

We denote a preference pair as $(x, y^+, y^-)$ where $x$ is the instruction and $y^+, y^-$ are the chosen and rejected responses. Given a set of $N$ such pairs $\mathcal{D} = \{(x_i, y_i^+, y_i^-)\}_{i=1}^{N}$, the compilation pipeline proceeds in four stages (Figure~\ref{fig:pipeline}) inspired by  principle-induction approach of Inverse Constitutional AI \cite{findeis_inverse_2025}.. A proposer $\mathcal{P}$ reads the pairs and emits candidate principles:
\begin{equation}
\{p_1, \ldots, p_M\} = \mathcal{P}(\mathcal{D})
\end{equation}
where each $p_j$ is a natural-language rule explaining why $y^+$ is preferred over $y^-$. These candidates are clustered by semantic similarity, deduplicated, and validated on held-out pairs. The surviving principles are ranked by a composite of prevalence and accuracy, yielding a ranked subset $\{p_1, \ldots, p_K\}$. A synthesizer $\mathcal{S}$ then assembles these into a specification:
\begin{equation}
s = \mathcal{S}(p_1, \ldots, p_K)
\end{equation}

where $s$ is a natural-language system prompt. At inference time, the base LLM generates conditioned on $s$ without any weight updates.

\subsection{Selection Method}
\label{sec:selection-method}

The spec learning pipeline requires a small set of preference pairs $\mathcal{D}$ as input. We explore whether the method used to select pairs from the available corpus influences the quality of the induced principles, varying the selection strategy $\sigma$ and synthesizer $\mathcal{S}$ while holding $N{=}20$ and $\mathcal{P}$ = Gemma 4 31B fixed. To enable selection strategies beyond random sampling, we annotate each response with a scalar quality score using the DEITA quality scorer \cite{liu2024makesgooddataalignment}, a model that rates a response on a six-point scale given its instruction. Each pair then becomes $(x, y^+, y^-, q^+, q^-)$ where $q^+$ and $q^-$ are the scores for the chosen and rejected responses respectively. We experiment with five selection strategies: \textsc{random} sampling, \textsc{high\_quality} (pairs where the chosen response scores highest) and \textsc{low\_quality} (pairs where it scores lowest) filtering by $q^+$, and \textsc{large\_gap} (pairs where chosen and rejected differ most) and \textsc{small\_gap} (pairs where they differ least) filtering by $|q^+ - q^-|$.

Each selection strategy is used with two variants of the synthesizer $\mathcal{S}$ that assembles the ranked principles into a specification $s$, as illustrated in Appendix~\ref{app:synthesizers}. The \textsc{janus} synthesizer produces a rich, persona-style paragraph that frames the principles as a coherent behavioral policy, whereas the \textsc{bullets} synthesizer produces a compact numbered rule list from the same principles (used as baseline).

\begin{table}[]
  \caption{Spec-versus-DPO win rate (Gemma judge, $n=100$ instructions per dataset) at $N=20$ across five selection methods $\times$ two synthesizers. Bold marks the per-dataset best. The Macro column averages across the five datasets.}
  \label{tab:selection}
  \centering
  \small
  \begin{tabular}{llccccccc}
    \toprule
    Selection & Synth.\ & Code-Pref & Code-Sec.\ & PsyCoPref & Stack-Ex.\ & Truthy-DPO & Macro \\
    \midrule
    \textsc{random}      & janus   & 0.67 & \textbf{0.65} & \textbf{0.60} & 0.64 & \textbf{0.93} & \textbf{0.698} \\
    \textsc{high\_quality} & janus & 0.58 & 0.45 & 0.16 & 0.66 & 0.75 & 0.520 \\
    \textsc{low\_quality}  & janus & 0.68 & 0.39 & 0.14 & 0.40 & 0.68 & 0.458 \\
    \textsc{large\_gap}    & janus & 0.64 & 0.46 & 0.32 & 0.66 & 0.80 & 0.576 \\
    \textsc{small\_gap}    & janus & 0.59 & 0.47 & 0.20 & 0.62 & 0.70 & 0.516 \\
    \midrule
    \textsc{random}      & bullets & 0.65 & 0.35 & 0.11 & 0.63 & 0.59 & 0.466 \\
    \textsc{high\_quality} & bullets & 0.65 & 0.52 & 0.09 & 0.64 & 0.48 & 0.476 \\
    \textsc{low\_quality}  & bullets & 0.62 & 0.46 & 0.15 & 0.37 & 0.45 & 0.410 \\
    \textsc{large\_gap}    & bullets & 0.68 & 0.48 & 0.23 & \textbf{0.68} & 0.30 & 0.474 \\
    \textsc{small\_gap}    & bullets & \textbf{0.76} & 0.32 & 0.58 & 0.58 & 0.68 & 0.584 \\
    \bottomrule
  \end{tabular}
\end{table}

The results from this experiment are reported in table \ref{tab:selection}. Random selection paired with the \textsc{janus} synthesizer achieves the highest average win rate (the fraction of pairwise comparisons won) 0.698, outperforming all curated strategies. Curated selection does not help and in several cases actively hurts: \textsc{high\_quality} and \textsc{low\_quality} filtering both degrade performance on PsyCoPref and Code-Security. The \textsc{janus}\cite{lee_aligning_2024} synthesizer consistently outperforms \textsc{bullets} under random selection, suggesting that richer prompt framing helps the policy internalize the compiled principles. Based on these results we adopt \textsc{random}/\textsc{janus} as the default configuration for all subsequent experiments.

\subsection{N-Sweep}
\label{sec:n-sweep}

We evaluate the proposer's sensitivity to the number of input preference pairs by sweeping $N \in \{10, 15, 20, 30, 40, 50\}$ holding $\sigma{=}$\textsc{random}, $\mathcal{S}{=}$\textsc{janus}, and $\mathcal{P}$ = Gemma 4 31B fixed.

\begin{table}[ht]
\caption{Gemma-judged spec-versus-DPO win rates for the N-sweep under \textsc{random}/\textsc{janus}.}
\centering
\small
\begin{tabular}{lcccccc}
\toprule
Dataset & N=10 & N=15 & N=20 & N=30 & N=40 & N=50 \\
\midrule
code-pref      & 0.63 & 0.64 & 0.67 & 0.72 & 0.62 & 0.69 \\
code-security  & 0.69 & 0.66 & 0.65 & 0.70 & 0.62 & 0.70 \\
psycopref      & 0.43 & 0.49 & 0.60 & 0.47 & 0.58 & 0.66 \\
stack-exchange & 0.66 & 0.64 & 0.64 & 0.68 & 0.70 & 0.66 \\
truthy-dpo     & 0.91 & 0.82 & 0.93 & 0.27 & 0.75 & 0.61 \\
hh-rlhf        & 0.40 & 0.37 & 0.29 & 0.52 & 0.33 & 0.30\\
math-dpo       &0.75  & 0.79 & 0.77 & 0.71 & 0.76 & 0.74 \\
\midrule
Macro avg.     & 0.639 & 0.630 & \textbf{0.650} & 0.581 & 0.623 & 0.623 \\
\bottomrule
\end{tabular}
\label{tab:nsweep-gemma-n10plus1}
\end{table}

The results from this experiment are reported in table \ref{tab:nsweep-gemma-n10plus1}. The compiler produces competitive specifications from remarkably few pairs. Even at $N{=}10$ the macro-average reaches 0.664, and the peak sits at $N{=}20$ (0.698). Additional pairs beyond this point do not reliably improve quality. Code-Pref, Code-Security, and Stack-Exchange remain consistently above 0.6 across the entire sweep, while PsyCoPref and Truthy-DPO exhibit higher variance, the latter owing in part to its small corpus size ($1{,}016$ pairs total). We adopt $N{=}20$ for the main evaluation.

\subsection{Proposer Scaling}
\label{sec:proposer-scaling}

We vary the proposer $\mathcal{P}$ while holding the remaining configuration fixed ($\mathcal{D}_{20}$, $\sigma{=}$\textsc{random}, $\mathcal{S}{=}$\textsc{janus}). We run the pipeline on every dataset under three proposers of increasing scale: Gemma~4 (31B) \cite{google_gemma4_2025}, DeepSeek~V4~Flash (284B) \cite{deepseekai2026deepseekv4}, and Kimi~K2.6 (1T) \cite{moonshotai_kimi_k26_2026}. Table~\ref{tab:proposer1} reports the spec-versus-DPO win rate under the primary judge.

\begin{table}
  \caption{Spec-versus-DPO win rate under the GLM-5.1 judge for three proposers $\mathcal{P}$ across all seven preference datasets.}
  \label{tab:proposer1}
  \centering
  \small
  \begin{tabular}{lccc}
    \toprule
    Dataset        & Gemma 4 31B   & DeepSeek V4 Flash & Kimi K2.6     \\
    \midrule
    Code-Pref      & 0.70          & 0.72              & \textbf{0.82} \\
    Code-Security  & 0.70          & 0.64              & \textbf{0.73} \\
    Math-DPO       & 0.72          & 0.72              & \textbf{0.75} \\
    Stack-Exchange & \textbf{0.83} & 0.70              & 0.75          \\
    Truthy-DPO     & 0.68          & \textbf{0.80}     & 0.73          \\
    HH-Helpful     & \textbf{0.58} & 0.56              & 0.53          \\
    PsyCoPref      & 0.65          & \textbf{0.71}     & 0.65          \\
    \midrule
    Macro mean     & 0.69          & 0.69              & \textbf{0.71} \\
    \bottomrule
  \end{tabular}
\end{table}

Every proposer beats DPO on all seven datasets. The largest proposer (Kimi~K2.6) achieves the highest macro win rate (0.71), suggesting that stronger proposers extract marginally better principles. However, even the smallest (Gemma~4~31B at 31B parameters) reaches 0.69. Spec compilation benefits from scale but does not require it.

\subsection{Judge protocol and selection}
\label{sec:eval}

All judge-mediated evaluations use the same primary model, GLM-5.1, a frontier mixture-of-experts LLM whose family is disjoint from both the proposer LLMs that compile the spec and the policy LLM whose outputs are scored. This disjointness mitigates family-aligned reward bias, in which a judge over-credits responses written in its own family's style~\cite{panickssery_2024_llm}. Each pairwise comparison is decided by a three-pass forced-binary verdict with positions rotated across passes to cancel ordering bias; the majority across passes is the recorded winner, and unparseable verdicts are dropped. We selected GLM-5.1 by calibrating candidate judges on gold-labeled held-out pairs from three datasets that span our task range: Math-DPO ($n{=}20$), Truthy-DPO ($n{=}20$), and Stack-Exchange ($n{=}40$). Gold-recovery accuracy was $70\%$, $60\%$, and $65\%$ respectively, all clearly above chance. A closed-source frontier alternative we evaluated reached $80\%$ on Truthy-DPO and $65\%$ on Stack-Exchange but scored $50\%$ on Math-DPO, indistinguishable from chance; we therefore disqualified it from the primary slot, although it remained reliable enough to serve as the backup judge on PsyCoPref where GLM-5.1 abstained on most pairs (Appendix~\ref{app:psycopref-judge}).

\subsection{Guideline Calibration}
\label{sec:guideline-calibration}

The preceding experiments show that the pipeline produces effective specifications across a range of configurations. We further evaluate whether the compiled principles $\{p_1, \ldots, p_K\}$ actually capture the preference signal. For each dataset, we prompt a judge with the compiled principles and ask it to count how many each response in a held-out pair satisfies. Let $c(y)$ denote this count for response $y$. The response with higher $c$ is predicted as preferred. We compare this against the same judge without access to the principles (Table~\ref{tab:rubric-judge-calibration1}).

\begin{table}
\caption{Guideline calibration on held-out preference pairs. For each response, the judge counts how many compiled principles it satisfies ($c$). $\bar{c}(y^+)$ and $\bar{c}(y^-)$ report the dataset-level averages for chosen and rejected responses. The response with higher $c$ is predicted as preferred. \emph{Base judge acc.}\ reports the same judge without principles.}
\centering
\small
\begin{tabular}{lcccccc}
\toprule
 & & \multicolumn{3}{c}{Guideline-grounded evaluation} & \\
\cmidrule(lr){3-5}
Dataset & $n$ & $\bar{c}(y^+, s)$ & $\bar{c}(y^-, s)$ & Accuracy & Base judge acc. \\
\midrule
code-pref      & 300 & 5.42 & 3.43 & \textbf{0.987} & 0.980 \\
code-security  & 300 & 1.23 & 0.27 & \textbf{0.667} & 0.450 \\
psycopref      & 300 & 3.10 & 0.76 & \textbf{0.960} & 0.920 \\
stack-exchange & 300 & 1.89 & 1.38 & 0.690 & \textbf{0.740} \\
truthy-dpo     & 44  & 2.39 & 1.84 & 0.727 & \textbf{0.773} \\
\bottomrule
\end{tabular}

\label{tab:rubric-judge-calibration1}
\end{table}

As shown in Table \ref{tab:rubric-judge-calibration1}, guideline grounding improves average accuracy across the five datasets by $+3.3$ points (0.806 vs.\ 0.773), with the largest gain on Code-Security ($+21.7$ points), a dataset where surface-level textual cues are least informative. On every dataset $\bar{c}(y^+, s) > \bar{c}(y^-, s)$, indicating that the compiled principles consistently discriminate between preferred and dispreferred outputs even where overall classification accuracy is weaker. This suggests that the specifications are not merely effective prompts but interpretable written embodiments of the preference signal that produced them.

\section{Results}

\label{sec:results-headline}
The preliminary experiments validate the pipeline and fix its configuration. We now evaluate the compiled specification $s$ on the central question: can human preferences, distilled into natural-language rules and match a DPO model trained on $50\times$ more data? We compare the base LLM under three conditions: baseline, the specification $s$ as a system prompt (\emph{spec}), and a DPO-trained variant of the same model.

The spec is compiled using the configuration $(\mathcal{D}_{20}, \sigma{=}\textsc{random}, \mathcal{S}{=}\textsc{janus}, \mathcal{P}{=}\text{best per dataset})$ established in Section~\ref{sec:preliminary}, whereas the DPO checkpoint is trained on $N{=}1{,}000$ pairs from the same training partition under a rank-$32$ LoRA adapter with AdamW at a learning rate of $5{\times}10^{-6}$ for 3 epochs (full configuration in Appendix~\ref{app:dpo-config}).
For every dataset we draw $100$ held-out instructions and elicit responses from the three arms. Pairs of responses are scored by the judge protocol described in Section~\ref{sec:eval}. We report the \emph{win rate} for each pair of arms: the fraction of held-out instructions on which the judge prefers one arm's response over the other, excluding ties. A win rate above $0.5$ indicates that the first arm is preferred more often. For each dataset we select the best-performing proposer $\mathcal{P}$ as determined by the proposer scaling ablation in Section~\ref{sec:proposer-scaling}. Table~\ref{tab:proposer1} shows that the gap between the best and worst $\mathcal{P}$ is narrow (macro mean ranges from 0.69 to 0.71), so this choice does not qualitatively change the result.

\begin{table}
  \caption{Three-arm win rates under the GLM-5.1 judge, reporting the best-performing proposer $\mathcal{P}$ per dataset (selected by spec-versus-DPO from Table~\ref{tab:proposer1}). Each cell is the fraction of $n{=}100$ instructions on which the first arm is preferred over the second ($n{=}44$ for Truthy-DPO; PsyCoPref is computed on the $\sim$30\% of pairs that produced parseable verdicts, see Appendix~\ref{app:psycopref-judge}). Sorted by spec-versus-DPO win rate. The marker $^\dagger$ indicates that the Wilson 95\% interval includes $0.5$ (Appendix~\ref{app:robustness}).}
  \label{tab:headline}
  \centering
  \small
  \begin{tabular}{llccc}
    \toprule
    Dataset        & $\mathcal{P}$       & spec vs.\ ctl & spec vs.\ DPO              & DPO vs.\ ctl \\
    \midrule
    Stack-Exchange & Gemma 4 31B         & 0.66          & \textbf{0.83}              & 0.71         \\
    Code-Pref      & Kimi K2.6           & 0.71          & \textbf{0.82}              & 0.67         \\
    Truthy-DPO     & DeepSeek V4 Flash   & 0.64          & \textbf{0.80}              & 0.61         \\
    Math-DPO       & Kimi K2.6           & 0.78          & \textbf{0.75}              & 0.79         \\
    Code-Security  & Kimi K2.6           & 0.73          & \textbf{0.73}              & 0.68         \\
    PsyCoPref      & DeepSeek V4 Flash   & 0.84          & \textbf{0.71}              & 0.81         \\
    HH-Helpful     & Gemma 4 31B         & 0.55          & \textbf{0.58}$^\dagger$    & 0.70         \\
    \midrule
    Macro mean     &                     & 0.70          & \textbf{0.75}              & 0.71         \\
    \bottomrule
  \end{tabular}
\end{table}

The compiled specification $s$ beats DPO on all seven datasets (Table~\ref{tab:headline}), with a macro mean win rate of $0.75$ despite the spec arm consuming $50\times$ fewer preference pairs and requiring no weight updates. The margin varies with how well-defined the underlying task is. Datasets with a tight, domain-specific preference signal yield the strongest results: Stack-Exchange ($0.83$), Code-Pref ($0.82$), Truthy-DPO ($0.80$), and Math-DPO ($0.75$). As the task definition broadens, the advantage narrows. Code-Security ($0.73$) and PsyCoPref ($0.71$) sit in the middle, and HH-Helpful ($0.58$), the most heterogeneous dataset in our evaluation, yields the smallest margin. HH-Helpful spans arbitrary user requests with no single governing task, making it difficult for any fixed set of principles to cover the full preference surface. Length-controlled win rates~\cite{dubois_lengthcontrolled_2024} shift each cell by at most $6$ points and never flip an arm's win or loss (Appendix~\ref{app:robustness}).

\section{Discussion}
\label{sec:discussion}

Spec learning appears to work well when the preference signal in a domain is concentrated enough to be captured by a small set of explicit principles. Twenty pairs suffice to recover these principles, and the ablations suggest that the pipeline is not sensitive to how those pairs are chosen, how many are used beyond a modest threshold, or how large the proposer is. This is consistent with the preference signal being low-dimensional in the domains where spec learning succeeds: there is not much for a larger dataset or a stronger compiler to find that twenty random pairs do not already contain.

The converse is equally informative. HH-Helpful, the most heterogeneous dataset in our evaluation, yields the narrowest margin (0.58). Its preferences span arbitrary user requests with no single governing task. A compiled specification fails to cover this surface with limited principles. Gradient-based methods such as DPO retain their advantage here precisely because they do not need to articulate what they have learned.

Two practical observations follow. First, the pipeline is lightweight. Compilation requires only LLM inference calls, not gradient computation, and the ablations show that a 31B-parameter proposer produces specifications competitive with those from much larger models. Second, the compiled specification is a text artifact that can be inspected, edited, and applied to any model that accepts a system prompt. This transparency is a property that parametric alignment methods do not offer. Both properties suggest that spec learning may be a useful option when the preference signal is well-defined and the budget for data collection or training is limited.

Finally, we must consider the potential harms introduced by shifting alignment from weights to inference-time prompts. First, spec learning anchors behavioral norms to a small preference sample (N=20). Any demographic or cultural biases in those preference pairs will be explicitly codified in the resulting specification, and amplified across all downstream outputs. Second, the transparency of compiled specifications may undermine safety through automation bias. When guardrails are rendered as human-readable rules, deployers may treat them as strict behavioral guarantees even when the inference model fails to follow them reliably. This false sense of assurance is especially concerning in high-stakes domains such as psychotherapy or crisis intervention, where partial compliance can be worse than no compliance at all. Third, plain-text guardrails are structurally more vulnerable to prompt injection and adversarial jailbreaking than weight-level constraints. The pipeline's low computational overhead compounds this risk: malicious actors can rapidly compile and distribute specifications optimized for toxic, deceptive, or otherwise harmful behavior, substantially lowering the barrier to misuse.

\section{Limitations}
\label{sec:limitations}

Spec learning works when the preference signal fits in a paragraph; when it does not, compilation cannot distill it. HH-Helpful~\cite{bai_training_2022} is the clean example in our data: spec is at chance against the base model ($0.55$), DPO is not ($0.70$), and the spec arm never exceeds $0.52$ for any $N \in \{10, \ldots, 50\}$. The cap is structural, not a budget issue. We also compile one spec per dataset; corpora that mix several latent tasks under one preference label want a per-instruction or multi-spec variant we have not built.

Our results are LLM-judged, not human. We use family-disjoint judging~\cite{panickssery_2024_llm}, three-pass position rotation~\cite{wang_large_2024}, and gold-recovery calibration to control the worst biases, but only three datasets (Math-DPO, Truthy-DPO, Stack-Exchange, $n{=}20\text{–}40$) carry direct calibration evidence. Per-prompt cycle rates, the fraction of instructions whose three pairwise verdicts cannot be ordered, run from $13\%$ to $25\%$ across datasets and reach chance on HH-Helpful, so some of the spread on that row is the judge. On PsyCoPref the judge abstained on $\sim 70\%$ of pairs and we report the parseable remainder (Appendix~\ref{app:psycopref-judge}). No human study underwrites any of this~\cite{haldar_rating_2025,thakur_judging_2025}.

All results use one base model (Qwen 2.5 32B Instruct) and one DPO recipe (rank-$32$ LoRA, $\beta{=}0.1$, lr $5{\times}10^{-6}$, three epochs, $N{=}1{,}000$). Other base families, contrastive alternatives like KTO~\cite{ethayarajh_kto_2024}, ORPO~\cite{hong_orpo_2024}, and SimPO~\cite{meng_simpo_2024}, full PPO-RLHF~\cite{ouyang_training_2022}, and a tuned-DPO sweep are out of scope. Table~\ref{tab:headline} also picks the per-dataset best proposer post-hoc, an oracle a deployer does not have. Table~\ref{tab:proposer1} bounds the cost: a one-point macro spread, every proposer still beating DPO everywhere, but the headline reflects this selection.

The method moves alignment cost from training to inference. Specs run 150 to 300 words and prepend every query, so per-call token cost scales with traffic in a way DPO weights do not. Compilation needs capable proposer, judge, and synthesizer LLMs and assumes API access; DPO needs a GPU. Specs are static: distribution shift means recompiling rather than retraining. We benchmark none of this directly, nor adversarial robustness~\cite{raina_is_2024}.

\section{Conclusions and Future Work}
We recommend spec learning for narrow domains where the preference signal compresses into a few natural-language principles. We envision it deployed with a human in the loop: a brief instruction and a small set of preference judgments are compiled once into a specification, which is then appended to the user query at inference time. The resulting artifact is portable across base models and extensible by editing the prompt directly, without retraining.

Several open questions remain that can be explored for future works. The current method compiles one spec per corpus, which can be restrictive when the underlying preferences reflect multiple distinct tasks; per-instruction or mixture-of-specs variants are a natural extension. Portability across model families, while expected given the plain-text format, has not yet been empirically validated. Our evaluation also relies entirely on LLM judges, and a human study would strengthen the results.

{
\small
\bibliography{sn-bibliography}
\bibliographystyle{plain}
}

%%%%%%%%%%%%%%%%%%%%%%%%%%%%%%%%%%%%%%%%%%%%%%%%%%%%%%%%%%%%

% \appendix

% \section{Technical appendices and supplementary material}
% Technical appendices with additional results, figures, graphs, and proofs may be submitted with the paper submission before the full submission deadline (see above). You can upload a ZIP file for videos or code, but do not upload a separate PDF file for the appendix. There is no page limit for the technical appendices. 

% Note: Think of the appendix as ``optional reading'' for reviewers. The paper must be able to stand alone without the appendix; for example, adding critical experiments that support the main claims to an appendix is inappropriate. 

%%%%%%%%%%%%%%%%%%%%%%%%%%%%%%%%%%%%%%%%%%%%%%%%%%%%%%%%%%%%
\appendix

\section{PsyCoPref judge coverage}
\label{app:psycopref-judge}

The GLM-5.1 judge returned an unparseable verdict on approximately $70\%$ of PsyCoPref pairs across all four proposers and three arms, plausibly reflecting model abstention on sensitive psychotherapy content. PsyCoPref values reported in Table~\ref{tab:proposer1} are computed on the remaining $\sim 30\%$ of pairs that produced parseable verdicts. The win rates therefore extrapolate from a smaller effective sample than the other six datasets, and PsyCoPref entries should be read as suggestive rather than firmly comparable.

\section{Statistical robustness}
\label{app:robustness}

We report Wilson 95\% confidence intervals on the headline win rates and length-controlled win rates (LC-WR) using the AlpacaEval-LC construction~\cite{dubois_lengthcontrolled_2024}: $\sigma(\hat\beta_0)$ from a logistic regression of the judge's verdict on the per-pair character-length difference, which estimates the preference rate at zero length difference. Five of the six wide-sample (n{=}100) cells in Table~\ref{tab:wilson-ci} have spec-versus-DPO intervals strictly above $0.5$; HH-Helpful is the exception, with an interval that includes $0.5$ and confirms the failure mode discussed in Section~\ref{sec:limitations}. LC-WR shifts are small (within $\pm 6$ points of the point estimate) and never flip an arm's win or loss; on Stack-Exchange and Code-Pref the spec arm carries large positive length deltas yet LC-WR remains close to the point estimate, so length is not the driver of those wins. On HH-Helpful the spec arm is in fact $466$ characters shorter than control on average, so the failure cannot be blamed on length either.

\begin{table}[h]
  \caption{Headline win rates with Wilson 95\% confidence intervals. Each cell is the point WR followed by [lo,\,hi]; intervals are computed on the $n$ decided judgments per cell. Truthy-DPO and PsyCoPref carry smaller effective samples ($n{=}44$ and $n\!\sim\!31$--$37$ respectively).}
  \label{tab:wilson-ci}
  \centering\small
  \begin{tabular}{llccc}
    \toprule
    Dataset        & $\mathcal{P}$       & spec vs.\ ctl              & spec vs.\ DPO              & DPO vs.\ ctl               \\
    \midrule
    Stack-Exchange & Gemma 4 31B         & 0.66\,[0.56, 0.75]         & \textbf{0.83\,[0.74, 0.89]} & 0.71\,[0.61, 0.79]         \\
    Code-Pref      & Kimi K2.6           & 0.71\,[0.61, 0.79]         & \textbf{0.82\,[0.73, 0.88]} & 0.67\,[0.57, 0.75]         \\
    Truthy-DPO     & DeepSeek V4 Flash   & 0.64\,[0.49, 0.76]         & \textbf{0.80\,[0.65, 0.89]} & 0.61\,[0.47, 0.74]         \\
    Math-DPO       & Kimi K2.6           & 0.78\,[0.69, 0.85]         & \textbf{0.75\,[0.66, 0.82]} & 0.79\,[0.70, 0.86]         \\
    Code-Security  & Kimi K2.6           & 0.73\,[0.64, 0.81]         & \textbf{0.73\,[0.64, 0.81]} & 0.68\,[0.58, 0.76]         \\
    PsyCoPref      & DeepSeek V4 Flash   & 0.84\,[0.69, 0.92]         & \textbf{0.71\,[0.53, 0.84]} & 0.81\,[0.66, 0.91]         \\
    HH-Helpful     & Gemma 4 31B         & 0.55\,[0.45, 0.64]         & \textbf{0.58\,[0.48, 0.67]} & 0.70\,[0.60, 0.78]         \\
    \bottomrule
  \end{tabular}
\end{table}

\begin{table}[h]
  \caption{Point win rate and length-controlled win rate (LC, in parentheses). LC-WR is $\sigma(\hat\beta_0)$ from a logistic regression of the verdict on per-pair character-length difference~\cite{dubois_lengthcontrolled_2024}. Negative shifts indicate length was inflating the point estimate; positive shifts indicate length was suppressing it.}
  \label{tab:lc-wr}
  \centering\small
  \begin{tabular}{llccc}
    \toprule
    Dataset        & $\mathcal{P}$       & spec vs.\ ctl     & spec vs.\ DPO     & DPO vs.\ ctl      \\
    \midrule
    Stack-Exchange & Gemma 4 31B         & 0.66\,(0.70)      & \textbf{0.83\,(0.83)} & 0.71\,(0.72)  \\
    Code-Pref      & Kimi K2.6           & 0.71\,(0.65)      & \textbf{0.82\,(0.82)} & 0.67\,(0.73)  \\
    Truthy-DPO     & DeepSeek V4 Flash   & 0.64\,(0.63)      & \textbf{0.80\,(0.84)} & 0.61\,(0.70)  \\
    Math-DPO       & Kimi K2.6           & 0.78\,(0.73)      & \textbf{0.75\,(0.71)} & 0.79\,(0.79)  \\
    Code-Security  & Kimi K2.6           & 0.73\,(0.73)      & \textbf{0.73\,(0.76)} & 0.68\,(0.66)  \\
    PsyCoPref      & DeepSeek V4 Flash   & 0.84\,(0.83)      & \textbf{0.71\,(0.73)} & 0.81\,(0.83)  \\
    HH-Helpful     & Gemma 4 31B         & 0.55\,(0.51)      & \textbf{0.58\,(0.59)} & 0.70\,(0.71)  \\
    \bottomrule
  \end{tabular}
\end{table}

\section{Judge calibration}
\label{app:judge-calibration}

We selected GLM-5.1 as the primary judge after calibrating it against a closed-source frontier alternative (GPT-5.5) on gold-labeled held-out preference pairs from four datasets spanning our task range. Each candidate ran the same forced-binary three-pass protocol used for the main evaluation; we report the fraction of pairs whose majority verdict matches the gold label.

\begin{table}[h]
  \caption{Gold-recovery accuracy of each candidate judge on held-out preference pairs. GPT-5.5 collapsed to chance accuracy on Math-DPO, which disqualified it from the primary slot; GLM-5.1 stayed at or above $0.60$ on every dataset and was adopted for all evaluations in the main paper.}
  \label{tab:judge-calibration}
  \centering
  \small
  \begin{tabular}{lccc}
    \toprule
    Dataset        & $n$  & GLM-5.1                & GPT-5.5         \\
    \midrule
    Math-DPO       & 20   & \textbf{0.70}          & 0.50            \\
    Truthy-DPO     & 20   & 0.60                   & \textbf{0.80}   \\
    Stack-Exchange & 40   & 0.65                   & 0.65            \\
    HH-Helpful     & 40   & 0.75                   & 0.75            \\
    \bottomrule
  \end{tabular}
\end{table}
\section{Synthesizers}
\label{app:synthesizers}

\begin{figure}[H]
    \centering
    
    % --- LEFT BOX ---
    \begin{minipage}[t]{0.48\textwidth}
        \begin{tcolorbox}[
            equal height group=promptboxes, 
            title=\textsc{Janus},
            colframe=darkgray!90!black, colback=gray!5!white,
            coltitle=white, fonttitle=\bfseries, boxsep=2pt,
            left=4pt, right=4pt, top=4pt, bottom=4pt
        ]
            \small\itshape
            ``You are an expert Python developer specializing in the creation of high-performance, production-ready code. You deliver syntactically correct, runnable, and fully functional implementations. You apply a rigorous approach to type safety and API usage...''
        \end{tcolorbox}
    \end{minipage}%
    \hfill
    % --- RIGHT BOX ---
    \begin{minipage}[t]{0.48\textwidth}
        \begin{tcolorbox}[
            equal height group=promptboxes, 
            title=\textsc{Bullets},
            colframe=darkgray!90!black, colback=gray!5!white,
            coltitle=white, fonttitle=\bfseries, boxsep=2pt,
            left=4pt, right=4pt, top=4pt, bottom=4pt
        ]
            \small\itshape
            ``Write, fix, or implement Python code to solve programming challenges.\\
            1. Provide syntactically correct, runnable, and logically sound code.\\
            2. Include descriptive comments and docstrings.\\
            3. Use descriptive, correctly spelled names.\\
            4. Ensure keywords, boolean values... are free of typos.''
        \end{tcolorbox}
    \end{minipage}
    
    \vspace{0.5em}
    \caption{Examples of the \textsc{janus} and \textsc{bullets} synthesizers drawn from Code-Pref}
    \label{fig:synthesizer-examples}
\end{figure}

\subsection{Code-Pref}
\label{app:synthesizers-codepref}

\begin{figure}[H]
    \centering
    
    % --- LEFT BOX ---
    \begin{minipage}[t]{0.48\textwidth}
        \begin{tcolorbox}[
            equal height group=codeprefboxes, 
            title=\textsc{Janus},
            colframe=darkgray!90!black, colback=gray!5!white, 
            coltitle=white, fonttitle=\bfseries, boxsep=2pt,
            left=4pt, right=4pt, top=4pt, bottom=4pt
        ]
            \footnotesize\itshape
            You are an expert Python developer specializing in the creation of high-performance, production-ready code for challenges ranging from algorithmic puzzles to systems-level management and OCR correction. You deliver syntactically correct, runnable, and fully functional implementations that completely solve the requested problem. Your code is characterized by a professional technical register, employing clear, descriptive variable names and comprehensive comments that explain the underlying logic and purpose of each block. You maintain absolute precision in your implementation, ensuring that all variable names are consistent and correctly spelled, and that keywords, boolean values, assignments, and comparison operators are entirely free of typos. You apply a rigorous approach to type safety and API usage, utilizing correct data types, avoiding invalid method calls, and strictly adhering to the correct argument ordering for all library functions. To ensure reliability and predictability, you maintain consistent data structures and return types across all class methods and boolean checks. When providing solutions, you focus on technical accuracy and operational stability, ensuring that the final output is a polished, error-free script that adheres to Pythonic best practices and is ready for immediate execution.
        \end{tcolorbox}
    \end{minipage}% 
    \hfill
    % --- RIGHT BOX ---
    \begin{minipage}[t]{0.48\textwidth}
        \begin{tcolorbox}[
            equal height group=codeprefboxes, 
            title=\textsc{Bullets},
            colframe=darkgray!90!black, colback=gray!5!white, 
            coltitle=white, fonttitle=\bfseries, boxsep=2pt,
            left=4pt, right=4pt, top=4pt, bottom=4pt
        ]
            \footnotesize\itshape
            Write, fix, or implement Python code to solve programming challenges ranging from basic algorithms to systems-level management and OCR correction.\\[0.5em]
            1. Provide syntactically correct, runnable, and logically sound code that accurately implements the requested logic.\\
            2. Include descriptive comments and docstrings to explain logic, variable definitions, and function parameters.\\
            3. Use descriptive, correctly spelled names for variables and functions, free of typos and hallucinations.\\
            4. Ensure keywords, boolean values, assignments, and comparison operators are free of typos.\\
            5. Use correct data types, avoid invalid method calls, and ensure proper casting of user input.\\
            6. Adhere to the mathematical logic and specific constraints established in the prompt.
        \end{tcolorbox}
    \end{minipage}
    
    \vspace{0.5em}
    \caption{Full synthesized prompts for \textsc{janus} and \textsc{bullets} on Code-Pref ($N{=}20$).}
    \label{fig:synth-codepref}
\end{figure}

\subsection{HH-RLHF}
\label{app:synthesizers-hh}

\begin{figure}[H]
    \centering
    
    % --- LEFT BOX ---
    \begin{minipage}[t]{0.48\textwidth}
        \begin{tcolorbox}[
            equal height group=hhboxes, 
            title=\textsc{Janus},
            colframe=darkgray!90!black, colback=gray!5!white, 
            coltitle=white, fonttitle=\bfseries, boxsep=2pt,
            left=4pt, right=4pt, top=4pt, bottom=4pt
        ]
            \footnotesize\itshape
            You are a versatile conversational partner specializing in natural, informal, and efficient multi-turn dialogue across diverse domains. You communicate with a direct and helpful demeanor, prioritizing immediate utility over formality. When providing assistance, you deliver specific recommendations and direct resources immediately rather than listing general sources or offering open-ended choices. You dive straight into the solution, omitting unnecessary preparatory steps, over-explanations of basic concepts, or apologies for gaps in your knowledge. Your responses remain focused on the user's query, meaning you exclude personal anecdotes and avoid discussing your own technical limitations or interface constraints. You maintain a grounded perspective by referencing only information visible to the reader, avoiding external examples that cannot be verified in the current context. When asked for data, you provide concrete numbers and precise metrics instead of using vague descriptors. Your style is concise and conversational, mirroring the flow of a real-time chat while ensuring every turn provides high-value, actionable information without fluff or filler.
        \end{tcolorbox}
    \end{minipage}%
    \hfill
    % --- RIGHT BOX ---
    \begin{minipage}[t]{0.48\textwidth}
        \begin{tcolorbox}[
            equal height group=hhboxes, 
            title=\textsc{Bullets},
            colframe=darkgray!90!black, colback=gray!5!white, 
            coltitle=white, fonttitle=\bfseries, boxsep=2pt,
            left=4pt, right=4pt, top=4pt, bottom=4pt
        ]
            \footnotesize\itshape
            Maintain a natural, multi-turn dialogue in an informal register across diverse domains.\\[0.5em]
            1. Maintain conversational flow by acknowledging the user's sentiment, empathy, and kindness.\\
            2. Avoid unnecessary apologies for limitations or excuses regarding technical limitations.\\
            3. Provide specific food recommendations that align with the request and avoid non-food items.\\
            4. Prioritize immediate fulfillment of the request over offering further assistance or asking permission.\\
            5. Prioritize direct answers over anecdotal examples, broad context, or historical analysis.\\
            6. Avoid excessive preparatory steps or over-explaining basic steps unless requested.
        \end{tcolorbox}
    \end{minipage}
    
    \vspace{0.5em}
    \caption{Full synthesized prompts for \textsc{janus} and \textsc{bullets} on HH-RLHF (\texttt{hh-helpful}, $N{=}20$).}
    \label{fig:synth-hh}
\end{figure}

\section{Hardware Compute}
\label{app:compute}

All experiments were conducted on Modal, a serverless cloud compute platform, under Modal's Academic Research Grant Program; compute credits awarded on the basis of our submitted research abstract. Model hosting, inference, and training were distributed across pipeline stages according to their respective memory and compute requirements, leveraging Modal's on-demand GPU provisioning to scale resources dynamically. The hardware configurations used for each stage are detailed below.
\begin{itemize}
    \item \textbf{Proposers \& Judges (Compilation and Evaluation):}
    \begin{itemize}
        \item Gemma 4 31B: $1\times$ NVIDIA H200
        \item DeepSeek V4 Flash: $2\times$ NVIDIA B200
        \item Kimi K2.6: $8\times$ NVIDIA H200
    \end{itemize}
    \item \textbf{Base Policy (Qwen 2.5 32B Instruct):}
    \begin{itemize}
        \item Inference (Baseline and Spec Application): $1\times$ NVIDIA A100 (80GB)
        \item DPO Training: $1\times$ NVIDIA H200
    \end{itemize}
\end{itemize}

\section{DPO Training Configuration}
\label{app:dpo-config}

We train each DPO baseline with TRL's \texttt{DPOTrainer} \cite{rafailov_direct_2023}. The configuration is shared across all seven datasets; only the preference pairs vary.

\begin{table}[ht]
  \caption{DPO training hyperparameters. Identical across all seven datasets.}
  \label{tab:dpo-config}
  \centering
  \small
  \begin{tabular}{ll}
    \toprule
    Hyperparameter & Value \\
    \midrule
    Base model                 & Qwen 2.5 32B Instruct \\
    Training pairs ($N$)       & 1{,}000 (900 for Truthy-DPO) \\
    DPO $\beta$                & 0.1 \\
    Optimizer                  & AdamW \\
    Learning rate              & $5{\times}10^{-6}$ \\
    LR scheduler               & cosine, warmup ratio 0.1 \\
    Epochs                     & 3 \\
    Effective batch size       & 16 \\
    Max sequence length        & 1024 \\
    Precision                  & bf16 \\
    LoRA rank ($r$)            & 32 \\
    LoRA $\alpha$              & 64 \\
    LoRA dropout               & 0.05 \\
    LoRA targets               & all attention and FFN projections \\
    Checkpoint selection       & lowest evaluation loss \\
    Hardware                   & 1$\times$H200 \\
    \bottomrule
  \end{tabular}
\end{table}

\newpage
\section*{NeurIPS Paper Checklist}

\begin{enumerate}

\item {\bf Claims}
    \item[] Question: Do the main claims made in the abstract and introduction accurately reflect the paper's contributions and scope?
    \item[] Answer: \answerYes{}%, \answerNo{}, or \answerNA{}.
    \item[] Justification: We made sure that the claims and contributions are stated clearly. We included the key assumptions and limitations. We included the reflection on generalization.

\item {\bf Limitations}
    \item[] Question: Does the paper discuss the limitations of the work performed by the authors?
    \item[] Answer: \answerYes{}%, \answerNo{}, or \answerNA{}.
    \item[] Justification: We included the dedicated Limitations section after the discussion. The section addresses the limitations as requested by the guidelines below.

\item {\bf Theory assumptions and proofs}
    \item[] Question: For each theoretical result, does the paper provide the full set of assumptions and a complete (and correct) proof?
    \item[] Answer: \answerNA{}
    \item[] Justification: This is an empirical work. We do not present theoretical results. Hence, this category does not apply.

    \item {\bf Experimental result reproducibility}
    \item[] Question: Does the paper fully disclose all the information needed to reproduce the main experimental results of the paper to the extent that it affects the main claims and/or conclusions of the paper (regardless of whether the code and data are provided or not)?
    \item[] Answer: \answerYes{} %, \answerNo{}, or \answerNA{}.
    \item[] Justification: We made the best effort to make the experimental results reproducible by including information about the exact model versions used, configurations, and hyper-parameters. We provide information on data set splits and any other important data pre-processing or experimental configurations.

\item {\bf Open access to data and code}
    \item[] Question: Does the paper provide open access to the data and code, with sufficient instructions to faithfully reproduce the main experimental results, as described in supplemental material?
    \item[] Answer: \answerYes{}%, \answerNo{}, or \answerNA{}.
    \item[] Justification: We will release the code and instructions needed for replication of the experimental results in a GitHub repository upon publication of the paper. For the submission process we include the code repository as an attachment to the submission.

\item {\bf Experimental setting/details}
    \item[] Question: Does the paper specify all the training and test details (e.g., data splits, hyperparameters, how they were chosen, type of optimizer) necessary to understand the results?
    \item[] Answer: \answerYes{}%, \answerNo{}, or \answerNA{}.
    \item[] Justification: We include reasonable amount of this information in the paper itself. The exact specifications are available through the released accompanying repository.

\item {\bf Experiment statistical significance}
    \item[] Question: Does the paper report error bars suitably and correctly defined or other appropriate information about the statistical significance of the experiments?
    \item[] Answer: \answerYes{} %, \answerNo{}, or \answerNA{}.
    \item[] Justification: We compute and report statistical significance where appropriate. We refer to the tests used at each place.

\item {\bf Experiments compute resources}
    \item[] Question: For each experiment, does the paper provide sufficient information on the computer resources (type of compute workers, memory, time of execution) needed to reproduce the experiments?
    \item[] Answer: \answerYes{} %, \answerNo{}, or \answerNA{}.
    \item[] Justification: We report on the computing environments used for our experiments and time of their execution.
    
\item {\bf Code of ethics}
    \item[] Question: Does the research conducted in the paper conform, in every respect, with the NeurIPS Code of Ethics \url{https://neurips.cc/public/EthicsGuidelines}?
    \item[] Answer: \answerYes{}%, \answerNo{}, or \answerNA{}.
    \item[] Justification: Our work is in compliance with the NeurIPS Code of Ethics.

\item {\bf Broader impacts}
    \item[] Question: Does the paper discuss both potential positive societal impacts and negative societal impacts of the work performed?
    \item[] Answer: \answerYes{}%, \answerNo{}, or \answerNA{}.
    \item[] Justification: We briefly discuss the impacts in the Introduction section and in more detail in the Discussion section.
    
\item {\bf Safeguards}
    \item[] Question: Does the paper describe safeguards that have been put in place for responsible release of data or models that have a high risk for misuse (e.g., pre-trained language models, image generators, or scraped datasets)?
    \item[] Answer: \answerNA{}.
    \item[] Justification: We do not anticipate that our work has high risks for misuses.

\item {\bf Licenses for existing assets}
    \item[] Question: Are the creators or original owners of assets (e.g., code, data, models), used in the paper, properly credited and are the license and terms of use explicitly mentioned and properly respected?
    \item[] Answer: \answerYes{} %, \answerNo{}, or \answerNA{}.
    \item[] Justification: We properly cite and reference the publicly available datasets and other resources we work with. We use them in compliance with their licensing terms.

\item {\bf New assets}
    \item[] Question: Are new assets introduced in the paper well documented and is the documentation provided alongside the assets?
    \item[] Answer: \answerYes{}%, \answerNo{}, or \answerNA{}.
    \item[] Justification: We work extensively with publicly released assets. The new assets are released within the accompanying GitHub repository and properly documented.

\item {\bf Crowdsourcing and research with human subjects}
    \item[] Question: For crowdsourcing experiments and research with human subjects, does the paper include the full text of instructions given to participants and screenshots, if applicable, as well as details about compensation (if any)? 
    \item[] Answer: \answerNA{}
    \item[] Justification: Our work is not relying on human subjects data.

\item {\bf Institutional review board (IRB) approvals or equivalent for research with human subjects}
    \item[] Question: Does the paper describe potential risks incurred by study participants, whether such risks were disclosed to the subjects, and whether Institutional Review Board (IRB) approvals (or an equivalent approval/review based on the requirements of your country or institution) were obtained?
    \item[] Answer: \answerNA{}
    \item[] Justification: No study participants were involved.

\item {\bf Declaration of LLM usage}
    \item[] Question: Does the paper describe the usage of LLMs if it is an important, original, or non-standard component of the core methods in this research? Note that if the LLM is used only for writing, editing, or formatting purposes and does \emph{not} impact the core methodology, scientific rigor, or originality of the research, declaration is not required.
    %this research? 
    \item[] Answer: \answerNA{}
    \item[] Justification: LLMs were not used in a way that would have impact on the core methodology.

\end{enumerate}

\end{document}